# A Survey on Computational Intelligence-based Transfer Learning

Mohamad Zamini, Eunjin Kim, *Member, IEEE*

*Abstract*—The goal of transfer learning (TL) is providing a framework for exploiting acquired knowledge from source to target data. Transfer learning approaches compared to traditional machine learning approaches are capable of modeling better data patterns from the current domain. However, vanilla TL needs performance improvements by using computational intelligence-based TL. This paper studies computational intelligence-based transfer learning techniques and categorizes them into neural network-based, evolutionary algorithm-based, swarm intelligence-based and fuzzy logic- based transfer learning.

*Index Terms*— computational intelligence, evolutionary algorithms, swarm intelligence, Transfer learning

## I. INTRODUCTION

TRADITIONAL machine learning approaches have been widely applied in different domains, and they reached state-of-the-art results in many problems. However, the common assumption of machine learning approaches to transfer learning is the source and target data have the same feature space or data distribution. Consequently, the model cannot perform well by changing the feature space or distribution of test data from the train data. Hence, the model must be retrained and rebuilt if the feature space changes, which is time consuming and expensive. In real-world problems, labeled data in many situations are not enough and also labeling new data can be impossible or expensive. Hence, transferring knowledge from one domain to another would be ideal for the industry.

In transfer learning knowledge from a source domain to a target domain will be transferred. Traditional Machine Learning (ML) approaches predictions are based on previously trained data on the same domain. But, transfer learning is able to have different domains in source and target data. The idea of transfer learning is inspired by the human brain which allows apply previously obtained knowledge as a solution for a new but similar problem. Researches in area of TL first proposed in 1995 with different names, including: learn to learn, meta learning, knowledge transfer, multi-task learning.  Transfer learning will be discussed in more detail in the following section.

The rest of the paper is structured as follows: Section II discusses transfer learning including its formal definition. Section III studies the usage of computational intelligence in transfer learning. Each topic of computational intelligence, including neural networks, evolutionary algorithms, swarm intelligence algorithms, and fuzzy logic, will be briefly reviewed in section III, followed by the conclusion of the paper in section IV. This study is among one of the first studies which have reviewed transfer learning using computational intelligence.

## II. TRANSFER LEARNING

Traditional machine learning approaches have been applied in different applications with state-of-the-art predictions. However, there are some limitations in some real-world applications. For example, in an ideal supervised learning scenario, a learning model requires a huge amount of labeled training data be collected which is time-consuming or even not possible. To solve such an issue, semi-supervised learning techniques were employed to help decrease the need of large-scale labeled data by just requiring few labeled data for a large-scale unlabeled data. In this scenario, it is sometimes impossible to collect large-scale unlabeled data, resulting in unsatisfactory results. Transfer learning can remove the requirement of huge amount of data by satisfying the hypothesis of data being independent and identically distributed with test set.

Before defining the transfer learning a domain and a task should be determined. According to [1], a feature space $X$ of domain $D$ and marginal distribution $P(X)$ where $D = \{X, P(X)\}$. In this definition X is an instance set defined as $X = \{x_i \mid x_i \in X, i = 1, ..., n\}$. A task $T$ consists of a label space $y$ and a decision function $f$, and is defined as $T = \{y, f\}$. The decision function $f$ learns from the sample data. A domain can be described by a couple of instances with or without a label [1]. Target domain is assumed to have some unlabeled instances and limited labeled instances. Transfer learning can be defined by a given observations related to $m^s \in \mathbb{N}^+$ in source domains $S_i$'s, and tasks $\{D_{S_i}, T_{S_i} \mid i = 1, ..., m^s\}$ and some observations where $m^T \in \mathbb{N}^+$ target domains $T_j$'s and tasks $\{D_{T_j}, T_{T_j} \mid j = 1, ..., m^T\}$.

The goal of transfer learning knowledge transfer across domains and can be a promising solution for the abovementioned problem. Learning to transfer can be

Mohamad Zamini and Eunjin Kim are with School of Electrical Engineering and Computer Science, University of North Dakota, Grand Forks, ND 58202 USA (e-mail: mohamad.zamini@ und.edu; ejkim@cs.und.edu).

concluded from experience generalization. Hence, a connection across learning activities of source and target domain is required. Transfer learning regarding consistency between source and target feature space and labels can be categorized into homogenous and heterogeneous transfer learning [2] (figure 1).

In homogenous transfer learning the assumption is they only differ in marginal distributions ($X^S = X^T$ and $y^S = y^T$), while in heterogeneous transfer learning the assumption will be in the situation which domains have different feature spaces ($X^S \neq X^T$ or/and $y^S \neq y^T$). Most homogenous transfer learning strategies can be categorized into either marginal distribution difference correction, conditional distribution difference correction or marginal and conditional distribution difference correction in source domain. Homogeneous TL can be used to propose a predictive model for the target domain where the same feature space with the source domain exists.

On the other hand, in heterogeneous transfer learning approaches it assumes source and target domain have similar distributions and the main focus is on alignment of source and target domain input spaces. Heterogeneous TL is still brand new topic. It can be used to unify or align the feature spaces or domains. In most current studies, it is assumed that source and target domain instances are from the same domain space.

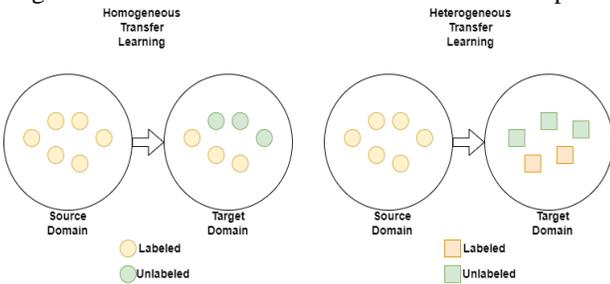

**Figure 1.** comparison of homogeneous vs heterogeneous transfer learning settings [3]

In homogeneous transfer learning studies, it is assumed that conditional distribution is the same in both domains and transfer can be done through instances with marginal distribution difference correction and use them in the target domain[4].

Transfer learning can also be through features of the source domain to match more closely with the target domain or by finding a common latent feature space by reducing the marginal distribution across source and target domain [5, 6]. The second TL approach is feature-based and is mostly used in domain adaptation learning. Domain adaptation is a setting of TL which transfers shared knowledge between different domains. In this setting they reuse knowledge extracted from source data in target data by directly reducing source and target distributions shifts through a proper feature representation.

Third approach of TL can be through shared parameters of both domains [7,[7, 8]. The knowledge at model/parameter level will be shared. For example, in object categorization, the object attributes can be transferred from source to target domains. The last approach of TL studies focused on defining some relations between both domains[9, 10]. It mainly focuses on the problems in the relational domains.

Early works in transfer learning applied AdaBoost [11] which is a statistical classification meta-algorithm. AdaBoost can be used with many other learning algorithms by increasing the next weak classifier accuracy with training instance weight adjustments. In the next section, we will discuss more how it has been combined with other learning algorithms.

## III. COMPUTATIONAL INTELLIGENCE IN TRANSFER LEARNING

Computational intelligence studies adaptive mechanisms to provide intelligent behavior in a complex environment using the computational models, including Artificial neural network (ANN) which is inspired by biological neural systems, fuzzy system (FS) which is inspired by the approximate human reasoning in the uncertain environment, evolutionary computation (EC) which inspired by natural evolution, swarm intelligence (SI) which inspired by the social behavior of living organisms which live in swarms or colonies, artificial immune systems (AIS) which is inspired by the human immune system, and probabilistic techniques [12]. Although machine learning approaches got considerable attention for transfer learning, few researchers focused on TR using evolutionary algorithms. In this section, we will discuss various computational intelligence models applied in transfer learning.

### A. Statistical meta-algorithms in transfer learning

One of the primary deep transfer learning (DTL) algorithms is TrAdaBoost [13] which selects those instances of the source domain that are similar to the target domain through re-weighted instances. In TrAdaBoost algorithm, a weak classifier on each iteration is constructed with the help of weighted instance data from the last iteration. In the next step, they increase the importance of misclassified instances in target domain and lowered the importance of misclassified instances in the source domain. TrAdaBoost on regression tasks cannot be well performed because the instances in the source domain can overcome the learning process.

TaskTrAdaBoost and MsTrAdaBoost are improved versions of TrAdaBoost [7]. Instead of single source knowledge transfer which will results in vulnerability to negative transfer, it imports knowledge from multiple sources to improve the chance of finding the closest source related to the target domain in both algorithms. MsTrAdaBoost is a multi-source TrAdaBoost, which extended TrAdaBoost by applying a transfer boosting algorithm to multiple source domains. The TaskTrAdaBoost also applies AdaBoost to each source domain to find candidate weak classifiers of them. The final selected classifier will minimize the target classification error. Then, at the next step, AdaBoost is applied to the labeled target data to adjust weights according to computed error. In [7], the authors applied an inductive TL and a popular boosting-based solution for object category recognition which reduces the negative knowledge transfer from multiple sources by promoting rapid retraining over the new target. The proposed model first exploits a traditional ML for extracting parameters that form the models of source task classifiers and sum up the knowledge from multiple sources. Then a parameter based approach boosts the target classifier. Both TaskTrAdaBoost and MsTrAdaBoost showed similar performance.

Similar to MsTrAdaBoost, MsTL-MvAdaboost is another instance-based DTL algorithms proposed by [14] that combines the concept of multi-source transfer learning and multi-view AdaBoost. However, this work does not use labeled data from several source tasks to improve learning of a single target task. Instead, this framework tries to automatically detect which parts of sample data are more specific for the source domain and which parts are more common across both domains. However, the proposed work has only been evaluated in a binary-class classification problem.

In the feature-based TL, most feature-based domain adaptation approaches tries to add adaptive modules in order to reduce distribution distance between source and target domains. Hence, an important challenge in domain adaptation is better understanding and proper using of distribution distance between domains. An early work by Pan et al. [5] proposed a shared latent space learning which minimizes the distribution distance between domains named maximum mean discrepancy embedding (MMDE). However, this method was not generalizable. As a result, authors proposed another algorithm named TCA.

A well-known feature-based DTL algorithm named transfer component analysis (TCA) proposed by Pan et al. [6]. This algorithm seeks to find a shared latent space for source and target distributions. TCA learns transfer components between domains in a Reproducing Kernel Hilbert Space (RKHS) using Maximum Mean Discrepancy (MMD) to train classifiers in the source domain to be used in the target domain. MMD loss computes the norm of the difference between two domain means. It minimizes the source and target feature distribution differences. It has been widely exploited and extended in traditional transfer learning. When the target domain is unlabeled, feature learning should minimize the feature distribution distance between domains. Hence, they proposed a semi-supervised TCA (SSTCA) to transform the learning kernel matrix problem to a low-rank matrix. In TCA, the sample means distance is measured in a lower- dimensional space which lacks the proper domain-specific features. Instead, the distance between distributions with MMD was measured and mapped it to the high dimensional RKHS in Baktashmotlagh's model [15]. The results compared to TCA shows superior performance.

In [16], Long, et.al. proposed another extension of MMD to measure the discrepancy of joint distributions, by using Hilbert space embeddings which is named Joint Maximum Mean Discrepancy (JMMD) and it improved their previous study [17]. JMMD applies non-uniform weights to better reflect the impact of variables in other layers and capture the interactions of them in the joint distributions.

*B. Transfer learning using neural networks*

The goal of neural networks is to solve complex nonlinear problems using human brain-inspired neural systems. Neural networks proved superior results in many real-world applications rather than statistical methods. Neural network-based TL algorithms have been recently widely studied. Deep neural networks are capable of learning transferable features for domain adaptation generalization. However, it is necessary to reduce dataset bias for better transferability in task-specific layers.

Unlike previous studies which used MMD with a focus on domain adaptation learning, Tzeng et al. in [18] studied transferring deep neural network representations between the source and target dataset with the help of optimizing the difference between source and target feature distributions. Similarly, Long et al. studied learning transferable features using a deep adaptation network with MMD for better generalizing to new tasks. In this approach they embedded task-specific layers into RKHS to better match the mean embeddings of domain distributions [19]. Multiple kernel variant MMD which is an improved version of MMD, is applied in deep adaptation network (DAN) architecture to generalize deep convolutional networks to adaptation scenarios. In this DAN, all hidden representations of different layers are embedded in a kernel Hilbert space to match domain distributions.

In [17], Long et al improved their model of [19], with jointly learning adaptive classifiers and transferable features from labeled source data to unlabeled target data. They used several layers to learn the residual function explicitly. Then the features are fused with the tensor product to perform lossless multi-layer feature fusion and embedded into RKHS to match the distributions for feature adaptation. The feature adaptation is performed by reducing MMD across domains. The proposed residual transfer network (RTN) can learn transferable features and adaptive classifiers jointly. The classifier adaptation is applied through connecting layers into deep networks to better learn residual function. The model assumes that source and target classifiers have a small residual function difference. The features of layers are combined with tensor product and embed them into kernel Hilbert spaces for feature adaptations. However, this approach does not consider the relationships between target samples and decision boundaries.

Beside MMD and its extensions, an adversarial loss is also used to minimize domain shifts. They learn a representation that can discriminate the source labels but not differentiate the domains. Adversarial learning methods are capable of improving recognition even with domain shift or dataset bias. In [20], Luo, et.al. proposed a framework to learn transferable representation across different domains and tasks efficiently. They used domain adversarial loss and a metric learning-based approach for the domain shift problem. In this approach, the assumption is data source is labeled and the target source is sparsely labeled or unlabeled.

A deep adversarial neural network (DANN) was proposed by [21] based on low-dimensional features shared between domains. The proposed domain classifier is inspired from generative adversarial network (GAN) to better learn transferable features adversarial. The discriminator in GANs is capable of determining if the instance is from the real data distribution or from the generated data by discriminating the real and fake distribution. In this work, the feature representation learner tries to learn indiscernible features from source and target domain. In the final stage a label classifier predicts the labels with the learned domain-invariant features.

Although the results are impressive, but it still lacks strong discriminative ability. Hence, domain-adversarial residual transfer (DART) learning model have been proposed which handles cross-domain image classification tasks [22]. This model introduced a perturbation function between the classifiers. They used ResNet as their source label classifier to learn perturbation function. This work focused on optimizing the joint distribution discrepancy of both image and labels rather than optimizing the discrepancy on marginal distributions in the adversarial learning scheme. The learning domain-invariant features of DART is through adversarial training.

A flexible generative adversarial domain adaptation network (G-DAN) with specific latent variables is proposed by [23]. The proposed model captures changes in the feature generation process between domains. G-DAN is also capable of generating data in new domains by setting new values for the latent variables. The model matches the feature distribution in the target domain to recover its joint distribution. They also introduced causal G-DAN (CG-DAN) which breaks down the joint distribution into different models to learn low-dimensional latent variables separately which can improve translation efficiency of the model.

Yoon et. al. [24] proposed multiple GAN architectures to learn transfer one dataset to another with the aim of enlarging the target dataset for learning better predictive models. This approach which is named RadialGAN addressed the feature mismatch and distribution mismatch by mapping all samples into the same latent space.

In [16], Tzeng, et.al. proposed adding a single fully connected layer to predict the domain label of inputs and proposed a domain confusion loss to lead the prediction to be close to a uniform distribution over binary label. Authors improved this work in [25] by proposing a novel generalized framework for adversarial adaptation by combining discriminative domain modeling, untied weight sharing, and a GAN loss. The proposed feature-based method is an unsupervised domain adaptation approach by mapping the target to the source feature space. This combined method is called Adversarial discriminative domain adaptation (ADDA), which is simpler than previous domain adversarial methods.

*C. Transfer learning using evolutionary algorithms*

Recently, a significant progress in prediction-based approaches enabled the integration of evolutionary algorithm (EA) and machine learning. Machine learning can derive a prediction model and the EA sustains the required performance with the environment changes over time. Evolutionary algorithms have been used in various ways.

To shift the initial population distribution of the target task to an optimized solution obtained from source tasks through genetic algorithm can be seen in [26] by Koçer, et.al. They proposed a transfer learning scheme for genetic algorithms. In this work, the knowledge from a multi-source problem is transferred to one target problem by copying some individuals of each generation into a pool. Then the available candidates were applied to replace some of the previously randomly generated individuals in the target domain. They aimed to produce a translation function using transfer learning to allow functions with differing values learned to be mapped from source to target tasks [27]. The results show that a transfer of inter-task mappings reduces the required learning time in a more complex task.

Another application of evolutionary algorithms in TL is in dynamic multi-objective optimization problems (DMOPs), which consist of multiple objective, constraint, and parameter that can change over time. The integration of TL approaches into an EA can improve the robustness and performance for designing more accurate dynamic multi-objective evolutionary algorithms. TL can reuse obtained knowledge from past experiences to improve the currently available solutions, which are computationally expensive. A memory-driven manifold TL-based evolutionary algorithm (MMTL-DMOEA/D) that combines memory mechanisms to maintain the best individuals from the past for optimal individual prediction and the manifold transfer learning feature to find the best solutions have been proposed in [28] by Jung, et.al. The proposed algorithm seeks for the best solution as initial population of the next generation. Results show the capability of better solutions with faster performance.

A solution for non-independent and identically distributed data in a dynamic environment is proposed by Jung [29]. The proposed TL-based dynamic multi-objective evolutionary algorithm (TrMOEA/D) system integrated TL and population-based evolutionary algorithms as a solution for DMOPs. This system reuses the past historical population after every change and generates an effective initial population pool by using past experiences to accelerate the evolutionary process. For this purpose, they adapted a TCA to construct a prediction model to gain knowledge of finding Pareto optimal solutions to generate an initial population pool for the optimization function of the next round. The proposed approach was applied in NSGA-II, MOPSO, and RM-MEDA evolutionary algorithms and experimented with twelve benchmark functions, showing impressive results.

To optimize the clustering accuracy by changing network structure and minimizing two successive clustering differences among two results in dynamic communities, Zou, et.al. proposed a feature transfer-based multi-objective optimization genetic algorithm (TMOGA) in [30]. This work extracts features from past community structures that are stable and also integrates this feature information into the current optimization process. They proposed feature transfer to solve the dynamic network problem. For feature transfer, they used a multi-objective evolutionary algorithm. The results show improvement in dynamic network community detection algorithms.

Despite successful applications of EAs in multi-objective problems, in majority of them all objectives can be evaluated at the same time. A surrogate assisted evolutionary algorithm (SAEA) combined with a parameter-based transfer learning is proposed by Wang [7] to transfer knowledge acquired by using common decision variables by adapting surrogate-assisted evolutionary algorithms to better deal with latencies. This approach applies a filter-based feature selection algorithm to better extract pivotal features of each objective. The proposed algorithm shows improvements compared to bi-objective optimization problems.

Another usage of evolutionary algorithms is in parameter selection in neural networks for transfer learning. PathNet in [8] is a parameter reuse neural network algorithm. It utilized neural networks with embedded agents to support transfer, continual, and multitask learning by determining which part of the neural network be reused for the new task. Agents are viewed through the network which seeks to find the subset of parameters used by forward and backward passes in the backpropagation algorithm. A tournament selection genetic algorithm (TSA) selects views (pathways) through the neural network for mutation. The proposed algorithm by Srivastava in [25] can be considered a form of evolutionary dropout in which dropout samples are evolved instead of randomly dropping out units. However, the proposed algorithm doesn't support more general networks.

Although PathNet chooses pre-trained modules in a modular neural network automatically, it still needs modular neural network as its pre-trained network. To make it more generalized by considering layers of a neural network as a modular network, Imai, et. al. in [31] proposed stepwise PathNet. In this model, a fixed-parameter or an adjustable-parameter layer is selected and TSA constructs the same architecture of the pre-trained neural networks [31]. However, these classifiers are also only trained all over the training process rather than the entire network.

The layer selection approach for TL of CNNs is still challenging. Using a genetic algorithm as a solution for layer selection has been proposed by Nagae, et.al. [32]. They proposed an efficient layer selection approach in which a genotype with high validation accuracy can be selected during genotype selection. A genotype represents which weights of layers are fixed or updated or fixed in TL. The experiments in the InceptionV3 network demonstrate more than 10% improvements compared to conventional methods for each selection method. In [33], de Lima Mendes, et.al. designed support vector regression predictor along with a multi-objective evolutionary algorithm based on decomposition (MOEA/D) for a dynamic multi-objective optimization problem. They used decomposition to convert the Pareto-optimal sets captured in successive environments into time series solutions. This algorithm selects trainable layers of InceptionV3 CNN for a classifier of Pneumonia images.

Instance weighting is a generic setting of statistical bias correction technique capable of sample correction. A new instance weighting framework for genetic programming-based symbolic regression is proposed by Chen, et.al. [34] to improve the cross-domain generalization ability. They applied differential evolution to find the optimal weights during genetic programming evolutionary process. It helps the model learn more important source domain features and ignores less useful source features. Compared with the three benchmark methods, the proposed method shows superior generalization performance on target sets. Despite its promising results in transfer learning genetic programming (TLGP), increasing ratio between the number of source and target domain instances will significantly increase, the computation cost of the proposed algorithm. To solve this issue, they proposed a new instance weighting framework in [35] by utilizing the density ratio estimation for transfer learning in GPSR to solve the limitations of the previous work. Thus, it shrinks the search space and provides better initial points for searching the optimal weights for source domain instances. Although the results were promising, one of the main aspects not considered is the distribution differences across source and target domains. Hence, it cannot be generalized in the scenarios that the source and target domain are very dissimilar.

In [36], Figueiredo et al. proposed GROOT, a framework that uses GA-based solutions to find the best mapping solution across the source and target domain. It works based on a set of relational dependency trees created from the source data to build for the target data. The results outperform baseline models on benchmark datasets.

The experiments of the relationships between genetic transfer and population diversification is studied by Gupta, et.al. [9]. This work, proposed transfer learning in multi-tasking optimization (MTO), which deals with multiple optimization tasks with various decision variables. Evolutionary multi-tasking as a novel model that enables similarities between distinct optimization tasks has been recently studied in [9]. Their multi-factorial evolutionary algorithm (MFEA) proposed in [10] exploits relationships between optimization tasks via multi-tasking through a cross-domain optimization platform. Although multitasking MFEA results in rapid streamlining of search towards feasible solutions in case of latent synergy existence, multi-tasking performance was the same as a single-tasking approach for the case of task clones.

To further improve MFEA, a two-level transfer learning (TLTL) algorithm has been proposed by Ma et al. [37] and inter-task transfer learning via chromosome crossover and elite individual learning is implemented in [34]. The intra-task of TL is based on information decision variables transfer for an across-dimension optimization. This solution shows outstanding results in global search and a fast convergence rate in TL. However, this method is similar to previous work that uses a prespecified scalar parameter *rmp* to organize the intensity of knowledge transfer. As a result, finding an optimal *rmp* without prior knowledge about the similarities between tasks to increase the efficiency of genetic algorithm is difficult, leading to performance degradation.

A new optimized label transfer approach to transform the observation between domains is proposed by Salaken et al. [38]. Unlike previous works, which used matching distribution between source and target domain, this work used learning domain invariant feature representations. Here, the assumption is both the transformed and the target domains have the same number of instances and observations. This label transfer from source data is performed by implementing a multivariate Gaussian mixture model (GMM). The genetic algorithm in this problem is applied to minimize the cost function and transformation process optimization which is a solution for distribution difference and improves the accuracy. The assumption of this work is the covariance matrix of the target data is positive and definite; otherwise, it will not work.

Stock markets have dynamic nature which requires agents to collaborate and transfer their best knowledge with each other. For this purpose, Hirchoua et al. applied transfer learning using the genetic algorithm to adapt the agent's internal strategies and force them to harmonize and transfer their experience with each other [39] . They used a learning method named a multi-agent reinforcement learning algorithm which does self-trades. The proposed model initiates by building an evolutive continuous training environment includes a bench of generations comprising a set of agents, and agents are able to collaborate and transfer knowledge.

A transfer learning-based EA has been proposed by Chugh et al. to match composite face sketch images to digital images [40]. They utilized a TL method in a genetic algorithm for composite sketch matching with digital images by a better learning matcher that increased the recognition performance. TL majorly helps in employing various sketch images accessible to the study matcher. HIM and HOG are the two feature extractors. The results on the IIT-D composite sketch with age variations (CSA) dataset show 34% accuracy for rank 10 of hand-drawn sketches and 5% for rank 1 of computer sketches.

Most evolutionary computation approaches start a search on the new problem of interest from scratch. A methodology to improve the evolutionary search on a new problem from previously solved problems is still challenging.

### D. Using swarm intelligence algortihms in transfer learning

Swarm Intelligence has recently attracted interest in various domains. Bonabeau defined swarm intelligence as "*The emergent collective intelligence of groups of simple agents*" in [41]. In other words, SI is the collective intelligence behaviour of self-organized and decentralized systems as it's defined by Ab Wahab [42]. SI is composed of two fundamental concepts: self-organization and labour division. Self-organization is the system's capability to evolve its agents to a suitable form [42]. When SI executes different tasks by individuals to make them capable of working together, it is called division of labour. Various swarm intelligence algorithms have been proposed, such as Particle swarm optimization (PSO), Ant colony optimization (ACO), Artificial bee colony optimization (ABC), Bacteria Foraging optimization (BFO), and others. To the best of our knowledge PSO has been applied to transfer learning. Therefore, we will mainly focus on PSO algorithm and its application in transfer learning in the following.

Many recent feature-based methods tried to propose a novel latent feature space in which projects both source and target domains. However, in most of these works, the dimensionality of latent feature space should be pre-defined. Instead of creating new feature space, feature selection for reducing differences between distributions has recently gotten more attention. Feature selection is not easy in large number of features and search space increases exponentially regarding that. On the other side, selecting an optimal feature subset is also hard and choosing a set of weakly relevant features can cause model performance to be degraded. To better feature selection, a global search method is needed. Most feature-based adaptation approaches make default assumptions about marginal distributions or conditional distributions for model simplicity. However, the ideal feature selection model is the one that considers preserving discriminative information, target variance maximization, marginal and conditional distribution divergence simultaneously, automatic parameter selection, and subspace divergence between source and target domains. Evolutionary algorithms have been widely studied in feature selection, and they have shown to be a good candidate. Swarm intelligence algorithms are getting more attention according to their simplicity and fewer parameters.

Particle swarm optimization is a swarm intelligence algorithm introduced by Eberhart and Kennedy, inspired by the behaviour of animals to force particles to search for global optimal solutions [43]. The studies in [44] [42] show that PSO has three key alignment factors: the behaviour of moving towards the average direction of local flock mates, separation which is for avoiding the crowded local flock mates, and cohesion which carries the algorithm towards the average position of local flock mates.

The PSO algorithm begins with the particles population initialization. Then, the fitness function, i.e., an objective function, computes the fitness values of those particles. In the subsequent steps, the personal best position of each particle and the global best position over the particles are getting updated, and the positions of the particles get further updated via their velocity updates. These processes iterate until the algorithm finds the global best particle position in the search space, meeting the termination condition.

Current TL models are not capable of minimizing marginal and conditional distribution, maximizing target domain variance, and modeling manifolds using geometric properties of source and target simultaneously. To overcome these drawbacks, PSO has been recently applied to TL. The early work was proposed by Nguyen et al. in 2018 [45]. They proposed a new feature-based TL method utilizing PSO. The new fitness function is applied leading PSO for automatic feature selection from original features and transfer source and target domains to a closer feature space. To keep the discriminative ability of selected features in both domains, the fitness function is designed concerning the k-nearest neighbor (KNN) classification to keep the number of model assumptions minimum and better accuracy. The results show competitive results with the four benchmark algorithms (TCA, STCA, MIDA, and SMIDA).

To increase the knowledge from source domain, the key factor is narrowing down the distribution differences between the source and target domain, as mentioned before. FSUTL-PSO by Sanodiya et al. is a feature selection unsupervised transfer learning based on particle swarm optimization approach with a novel fitness function which integrates all essential goals [46]. It uses the PSO technique for the selection of proper feature subset from source and target. As a result, the evaluations on Office+Caltech and PIE Face datasets compared to other TL and non-TL methods shows superiority.

UDATL-PSO by Sanodiya et al. is an unsupervised discriminant analysis for transfer learning framework that considers all the quality measures simultaneously[47]. They used PSO to select the best values of different parameters in a common framework by using a nature-inspired computing of SI. A new fitness function that alleviate the discrepancy between domains and selecting a proper parameter values automatically with the help of the PSO-based parameter selection algorithm is the novelty of this work. To do this, they used linear discriminant analysis for considering source domain discriminative information, followed by PCA and LDA for increasing the target domain variance depending on the size of labeled data and utilizing a Laplacian regularization on fitness function to manipulate the manifold geometrical attributes of nearest points of source and target domains. Finally, they used maximum mean discrepancy for diminishing marginal distribution between domains. The results in Office+Caltech and PIE Face datasets have been experimented with.

Gomes et al. proposed a parameter selection approach for SVM [48]. For this purpose, they combined meta-learning and search algorithms. They utilized meta-learning as SVM parameter values recommender based on parameter configurations that showed successful results in similar problems stored in Meta-Database (MDB). The returned parameter values by meta-learning are then applied in the initial search points by a search technique for further parameter selection. The results showed the initial solution by solely meta-learning is located in promising search space regions. In the search module, new parameter candidates will be generated iteratively. The combination of meta-learning with PSO and Tabu search algorithm showed it could obtain lower error rates. This work is a mature version of [49] which they just applied PSO as a search algorithm to choose the values of two SVM parameters for regression problems. Although it showed good results it seems using a single objective search technique is not a good candidate for SVM parameter selection because it is a multi-objective optimization problem.

Another similar work for the SVM parameter selection problem using PSO was proposed by Miranda et al. [50]. As mentioned before, the SVM learning process is an inherently multi-objective optimization. Therefore, they proposed a multi-objective PSO to maximize the success rate and minimize the number of support vectors of the model based on well-succeeded parameters adopted in previous similar problems. This work also used MOPSO to find the values of two SVM parameters for classifications.

To improve the limitations of traditional evolutionary algorithms like a genetic algorithm which are mostly used in solution quality and convergence speed, an ant colony optimization (FEACO) has been proposed by [51] to improve the speed and efficiency. In this study, they applied a new pheromone updating technique to the original ant colony optimisation algorithm. Additionally, to solve the colored traveling salesman problem (CTSP), multi-task cooperative learning to cooperate ants to find the best solutions by using pheromone has been applied. In the proposed model, they replace the drift operator of the ITÖ algorithm instead of pheromone updating rule of ant colony optimization. This algorithm is effectively applied on multiple mechanisms like PSO and simulated annealing process and showed its drift operator performs well in finding a good solution. This study also needs to be evaluated under large-scale data.

Ghasemi Darehnaei et al. [52] proposed a swarm intelligence ensemble deep TL (SI-EDTL) for multiple vehicle detection on a UAV dataset to determine the regions into different vehicle types (car, truck, van, and bus). They used three R-CNNs (InceptionV3, ResNet50, and GoogleNet) combined with five transfer classifiers using a weighted averaging aggregation. The hyperparameters of the proposed model are tuned by applying a whale optimization algorithm which has shown to be superior to existing techniques.

*E. Transfer learning using fuzzy systems*

In many real-world applications, uncertainty is evident and applying fuzzy logic in TL can help us get better performance. In this section, the concept of combining transfer learning and fuzzy logic will be discussed.

Takagi-Sugeno-Kang fuzzy system (TSK-FS) has been widely applied in transfer learning approaches. Deng et al. proposed to use the development of transfer learning TSK FS named the knowledge leverage-based TSK FS (KL-TSK-FS) [53]. In this model, knowledge of the source domain can complete the insufficient data in the target domain. Compared with knowledge leverage-based Mamdani-Larsen fuzzy system (KL-ML-FS) which is a kernel density estimation-based TL[54], KL-TSK-FS has better flexibility because of the steadiness of its transfer learning mechanism. However, the knowledge transfer mechanism of KL-TSK-FS from source to target is not suitable for learning of consequents. Hence, Deng et al. enhanced KL-TSK-FS with two knowledge leverage strategies of learning both antecedent parameters and consequent parameter[55]. The results show better performance compared to previous works. However, the proposed transfer learning mechanism still needs a more adaptive TSK FS modeling like type 2 fuzzy systems.

Unlike previous works which were inductive, Yang et al. applied Takagi-Sugeno-Kang (TSK) fuzzy logic system based on transductive transfer learning [56]. In this model, the TSK fuzzy logic system is often used to prevent the drift distribution of data between the source and target domains. For this purpose, they proposed two objective functions to train a TSK fuzzy logic system. The proposed objective function is applied for regression (TSK-TL-FLS Reg) and binary classification (TSK-TL-FLS BC). The data used for the experiment was EEG signals, a multi-class classification problem. The results show superiority to the classic methods like LDA, SVM, decision tree, fuzzy logic system, and naïve Bayes.

The other study of TSK in transfer learning can be seen in [57]. Chen et al. studied ResTL for regression problems where the source and target domains share the same marginal distribution but have different conditional probability distributions. They applied TSK fuzzy system to source model parameters and reused them. However, the TSK by itself couldn't be constructed by few labeled target data. Hence, they proposed to reuse the antecedent parameters of the source fuzzy system and the target model obtained with ResTL which is the sum of fuzzy residual and model agnostic (source model). The problem with this approach is that increasing the size of the target data doesn't improve the results satisfactory enough.

Fuzzy transfer learning (FTL) was developed to deal with uncertainty. However, traditional FTL is unable to resolve multi-source knowledge domain combinations, so it was used for the source domains. A new method by Lu et al. proposed to combine fuzzy rules from multiple domains for regression tasks [57]. The results in homogenous and heterogeneous spaces are impressive.

A study focusing on fuzzy regression transfer learning has been proposed by Zuo et al. to address the value of the target for regression estimation [58]. In this model a Takagi-Sugeno regression model is applied to transfer knowledge from source to target domain. The proposed model performed superior to previous regression models.

Shell and Coupland proposed fuzzyTL to address learning tasks that have no prior direct contextual knowledge [59]. Their framework learns target tasks using a fuzzy inference system using limited unlabeled target data and labeled source data. In the framework no or little unlabeled data is available. The framework uses an adaptive online learning methodology to improve FIS transfer among contextually different learning tasks.

## IV. CONCLUSION

In this paper we reviewed currently available literatures on how computational intelligence techniques can help transfer learning. Our focus was given to applying four main groups of computational intelligence to TL: the neural network-based TL, evolutionary algorithms-based TL, swarm intelligence algorithms-based TL, and fuzzy logic-based TL. As we studied in this paper, while neural network and deep learning with deep NN have been widely studied in transfer learning, TL applying fuzzy logic, evolutionary algorithms, and swarm intelligence-based techniques are emerging research topic in this area.

In many current studies of transfer learning, source and target domain are assumed to be the same. In the future work, it needs to focus on how computational intelligence approaches can be further extended to transfer knowledge between different domains. In addition, it is also considered required to study more details about the computational complexity of CI-based TL models and how to develop a new model whose complexity is reduced and further optimized.

## REFERENCES

[2] Zhuang, F., et al., A comprehensive survey on transfer learning. Proceedings of the IEEE, 2020. 109(1): p. 43-76.
[3] Weiss, K., T.M. Khoshgoftaar, and D. Wang, A survey of transfer learning. Journal of Big data, 2016. 3(1): p. 1-40.
[4] Mignone, P., et al., Multi-task learning for the simultaneous reconstruction of the human and mouse gene regulatory networks. Scientific Reports, 2020. 10(1): p. 1-15.
[5] Pan, S.J., J.T. Kwok, and Q. Yang. Transfer learning via dimensionality reduction. in AAAI. 2008.
[6] Pan, S.J., et al., Domain adaptation via transfer component analysis. IEEE transactions on neural networks, 2010. 22(2): p. 199-210.
[7] Wang, X., et al. Transfer learning for gaussian process assisted evolutionary bi-objective optimization for objectives with different evaluation times. in Proceedings of the 2020 genetic and evolutionary computation conference. 2020.
[8] Fernando, C., et al., Pathnet: Evolution channels gradient descent in super neural networks. arXiv preprint arXiv:1701.08734, 2017.
[9] Gupta, A. and Y.-S. Ong. Genetic transfer or population diversification? Deciphering the secret ingredients of evolutionary multitask optimization. in 2016 IEEE Symposium Series on Computational Intelligence (SSCI). 2016. IEEE.
[10] Gupta, A., Y.-S. Ong, and L. Feng, Multifactorial evolution: toward evolutionary multitasking. IEEE Transactions on Evolutionary Computation, 2015. 20(3): p. 343-357.
[11] Freund, Y. and R.E. Schapire, A decision-theoretic generalization of on-line learning and an application to boosting. Journal of computer and system sciences, 1997. 55(1): p. 119-139.
[12] Andries, P., Computational intelligence: an introduction. 2022.
[13] Dai, W., et al., Boosting for transfer learning, in Proceedings of the 24th international conference on Machine learning. 2007, Association for Computing Machinery: Corvalis, Oregon, USA. p. 193–200.
[14] Xu, Z. and S. Sun. Multi-source transfer learning with multi-view adaboost. in International conference on neural information processing. 2012. Springer.
[15] Baktashmotlagh, M., et al. Unsupervised domain adaptation by domain invariant projection. in Proceedings of the IEEE International Conference on Computer Vision. 2013.
[16] Long, M., et al. Deep transfer learning with joint adaptation networks. in International conference on machine learning. 2017. PMLR.
[17] Long, M., et al., Unsupervised domain adaptation with residual transfer networks. Advances in neural information processing systems, 2016. 29.
[18] Tzeng, E., et al., Deep domain confusion: Maximizing for domain invariance. arXiv preprint arXiv:1412.3474, 2014.
[19] Long, M., et al. Learning transferable features with deep adaptation networks. in International conference on machine learning. 2015. PMLR.
[20] Luo, Z., et al., Label efficient learning of transferable representations acrosss domains and tasks. Advances in neural information processing systems, 2017. 30.
[21] Ganin, Y., et al., Domain-adversarial training of neural networks. The journal of machine learning research, 2016. 17(1): p. 2096-2030.
[22] Fang, X., et al., DART: domain-adversarial residual-transfer networks for unsupervised cross-domain image classification. Neural Networks, 2020. 127: p. 182-192.
[23] Gong, M., et al., Causal generative domain adaptation networks. arXiv preprint arXiv:1804.04333, 2018.
[24] Yoon, J., J. Jordon, and M. Schaar. RadialGAN: Leveraging multiple datasets to improve target-specific predictive models using Generative Adversarial Networks. in International Conference on Machine Learning. 2018. PMLR.
[25] Tzeng, E., et al. Adversarial discriminative domain adaptation. in Proceedings of the IEEE conference on computer vision and pattern recognition. 2017.
[26] Koçer, B. and A. Arslan, Genetic transfer learning. Expert Systems with Applications, 2010. 37(10): p. 6997-7002.
[27] Lu, J., et al., Transfer learning using computational intelligence: A survey. Knowledge-Based Systems, 2015. 80: p. 14-23.
[28] Jiang, M., et al., A fast dynamic evolutionary multiobjective algorithm via manifold transfer learning. IEEE Transactions on Cybernetics, 2020. 51(7): p. 3417-3428.
[29] Jiang, M., et al., Transfer learning-based dynamic multiobjective optimization algorithms. IEEE Transactions on Evolutionary Computation, 2017. 22(4): p. 501-514.
[30] Zou, J., et al., Transfer Learning Based Multi-Objective Genetic Algorithm for Dynamic Community Detection. arXiv preprint arXiv:2109.15136, 2021.
[31] Imai, S., S. Kawai, and H. Nobuhara, Stepwise pathnet: a layer-by-layer knowledge-selection-based transfer learning algorithm. Scientific Reports, 2020. 10(1): p. 1-14.
[32] Nagae, S., S. Kawai, and H. Nobuhara. Transfer learning layer selection using genetic algorithm. in 2020 IEEE Congress on Evolutionary Computation (CEC). 2020. IEEE.
[33] de Lima Mendes, R., et al. Many Layer Transfer Learning Genetic Algorithm (MLTLGA): a New Evolutionary Transfer Learning Approach

Applied To Pneumonia Classification. in 2021 IEEE Congress on Evolutionary Computation (CEC). 2021. IEEE.

[34] Chen, Q., B. Xue, and M. Zhang. Differential evolution for instance based transfer learning in genetic programming for symbolic regression. in Proceedings of the genetic and evolutionary computation conference companion. 2019.

[35] Chen, Q., B. Xue, and M. Zhang, Genetic programming for instance transfer learning in symbolic regression. IEEE Transactions on Cybernetics, 2020.

[36] Figueiredo, L.F.d., A. Paes, and G. Zaverucha. Transfer Learning for Boosted Relational Dependency Networks Through Genetic Algorithm. in International Conference on Inductive Logic Programming. 2021. Springer.

[37] Ma, X., et al., A two-level transfer learning algorithm for evolutionary multitasking. Frontiers in Neuroscience, 2020: p. 1408.

[38] Salaken, S.M., et al. Semi-Supervised Transfer Learning with Genetic Algorithm Tuned Transformation and Novel Label Transfer Mechanism. in 2018 IEEE International Conference on Systems, Man, and Cybernetics (SMC). 2018. IEEE.

[39] Hirchoua, B., et al. Evolutionary Deep Reinforcement Learning Environment: Transfer Learning-Based Genetic Algorithm. in The 23rd International Conference on Information Integration and Web Intelligence. 2021.

[40] Chugh, T., et al. Transfer learning based evolutionary algorithm for composite face sketch recognition. in Proceedings of the IEEE Conference on Computer Vision and Pattern Recognition Workshops. 2017.

[41] Bonabeau, E., et al., Swarm intelligence: from natural to artificial systems. 1999: Oxford university press.

[42] 42. Ab Wahab, M.N., S. Nefti-Meziani, and A. Atyabi, A comprehensive review of swarm optimization algorithms. PloS one, 2015. 10(5): p. e0122827.

[43] Eberhart, R. and J. Kennedy. Particle swarm optimization. in Proceedings of the IEEE international conference on neural networks. 1995. Citeseer.

[44] Del Valle, Y., et al., Particle swarm optimization: basic concepts, variants and applications in power systems. IEEE Transactions on evolutionary computation, 2008. 12(2): p. 171-195.

[45] Nguyen, B.H., B. Xue, and P. Andreae, A particle swarm optimization based feature selection approach to transfer learning in classification, in Proceedings of the Genetic and Evolutionary Computation Conference. 2018, Association for Computing Machinery: Kyoto, Japan. p. 37–44.

[46] Sanodiya, R.K., et al., A particle swarm optimization-based feature selection for unsupervised transfer learning. Soft Computing, 2020. 24(24): p. 18713-18731.

[47] Sanodiya, R.K., et al., Particle swarm optimization based parameter selection technique for unsupervised discriminant analysis in transfer learning framework. Applied Intelligence, 2020. 50(10): p. 3071-3089.

[48] Gomes, T.A., et al., Combining meta-learning and search techniques to select parameters for support vector machines. Neurocomputing, 2012. 75(1): p. 3-13.

[49] Gomes, T.A., et al. Combining meta-learning and search techniques to svm parameter selection. in 2010 Eleventh Brazilian Symposium on Neural Networks. 2010. IEEE.

[50] Miranda, P.B., et al. Multi-objective optimization and Meta-learning for SVM parameter selection. in The 2012 International Joint Conference on Neural Networks (IJCNN). 2012. IEEE.

[51] Dong, X., W. Dong, and Y. Cai, Ant colony optimisation for coloured travelling salesman problem by multi-task learning. IET Intelligent Transport Systems, 2018. 12(8): p. 774-782.

[52] Ghasemi Darehnaei, Z., et al., SI-EDTL: Swarm intelligence ensemble deep transfer learning for multiple vehicle detection in UAV images. Concurrency and Computation: Practice and Experience, 2022: p. e6726.

[53] Deng, Z., et al., Knowledge-leverage-based TSK fuzzy system modeling. IEEE transactions on neural networks and learning systems, 2013. 24(8): p. 1200-1212.

[54] Deng, Z., et al., Knowledge-leverage-based fuzzy system and its modeling. IEEE Transactions on Fuzzy Systems, 2012. 21(4): p. 597-609.

[55] Deng, Z., et al., Enhanced Knowledge-Leverage-Based TSK Fuzzy System Modeling for Inductive Transfer Learning. ACM Trans. Intell. Syst. Technol., 2016. 8(1): p. Article 11.

[56] Yang, C., et al., Takagi–Sugeno–Kang transfer learning fuzzy logic system for the adaptive recognition of epileptic electroencephalogram signals. IEEE Transactions on Fuzzy Systems, 2015. 24(5): p. 1079-1094.

[57] Chen, G., Y. Li, and X. Liu, Transfer learning under conditional shift based on fuzzy residual. IEEE Transactions on Cybernetics, 2020.

[58] Zuo, H., et al., Fuzzy regression transfer learning in Takagi–Sugeno fuzzy models. IEEE Transactions on Fuzzy Systems, 2016. 25(6): p. 1795-1807.

[59] Shell, J. and S. Coupland, Fuzzy transfer learning: methodology and application. Information Sciences, 2015. 293: p. 59-79.

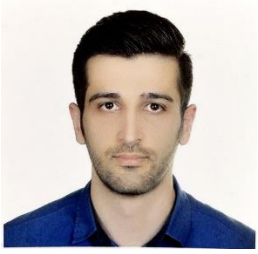
**Mohamad Zamini** I am a PhD student at the Department of Computer Science at University of North Dakota supervised by Prof. Hassan Reza. I received my master degree from the department of information technology, Tarbiat Modares University, Iran, in 2018 supervised by Prof. Gholamali Montazer. Prior to that, I received my bachelor degree in software engineering from University of science and culture in 2016. I am interested in data mining, machine learning, and natural language processing.

**Eunjin Kim**, She is an associate professor at the Department of Computer Science at University of North Dakota.